\RequirePackage{fix-cm}
\documentclass[smallextended]{svjour3}       
\smartqed  
\usepackage{graphicx}
\usepackage{amsmath}
\usepackage{amssymb}
\usepackage{algorithm}
\usepackage{algpseudocode}
\usepackage{color}
\usepackage{ulem}

\begin{document}

\title{Object Detection and Tracking Benchmark in Industry Based on Improved Correlation Filter}


\author{Shangzhen Luan    \and
        Yan Li \and
        Xiaodi Wang\and
        Baochang Zhang*
}


\institute{Shangzhen Luan, Xiaodi Wang and Baochang Zhang(Correspondence, bczhang@139.com)\at
           School of Automation Science and Electrical Engineering, Beihang University, Beijing,China 
           \and
           Yan Li  \at
             School of Electronics and Information Engineering, Beihang University, Beijing China
           \and
           Baochang Zhang is also with Shenzhen Academy of Aerospace Technology, Shenzhen, China.              
}

\date{Received: 31 August 2017 / Revised: 28 March 2018 / Accepted: 30 April 2018\\
© Springer Science+Business Media, LLC, part of Springer Nature 2018}

\maketitle

\begin{abstract}
Real-time object detection and tracking have shown to be the basis of intelligent production for industrial 4.0 applications. It is a challenging task because of various distorted data in complex industrial setting. The correlation filter (CF) has been used to trade off the low-cost computation and high performance. However, traditional CF　training strategy can not get satisfied performance for the various industrial data; because the simple sampling(bagging) during training process will not find the exact solutions in a data space with a large diversity. In this paper, we propose Dijkstra-distance based correlation filters (DBCF), which establishes a new learning framework that embeds distribution-related constraints into the multi-channel correlation filters (MCCF). DBCF is able to handle the huge variations existing in the industrial data by improving those constraints based on the shortest path among all solutions. To evaluate DBCF, we build a new dataset as the benchmark for industrial 4.0 application. Extensive experiments demonstrate that DBCF produces high performance and exceeds the state-of-the-art methods.{\color{blue}The dataset and source code can be found at https://github.com/bczhangbczhang.}
\end{abstract}

\section{Introduction}
\label{intro}
Many industrial applications require detection and tracking of various objects in real-time\cite{2008Real}, which is a fundamental task of intelligent production and smart environment\cite{Han2012Employing} for industrial 4.0 applications. In this setting, a simple and highly efficient method is the key for trade off low-computation and robustness during separating various objects from cluttered background. For this purpose, we choose a useful tool, correlation filter, which is usually trained in the frequency domain with the aim of producing a strong correlation peak on the pattern of interest while suppressing the response to the background.

The correlation filters were first proposed by Hester and Casasent, and named synthetic discriminant functions (SDF) \cite{[9]}, which focus more on formulating the theory. To facilitate more practical applications, many variants were proposed including both constrained (such as \cite{[10],[11]}) and unconstrained correlation filters \cite{[12],[13]}. In recent years, methods based on correlation filters have been widely developed on object detection and tracking. For object detection tasks, the average of synthetic exact filters (ASEF)\cite{[1]} was successfully applied to eye localization. Instead of using intensity feature, multi-channel correlation filters (MCCF) \cite{[3],[15]} took advantages of multi-channel features, such as histogram of oriented gradients (HOG)\cite{[4]} to enhance the performance of classifier. Vector correlation filters (VCFs)\cite{[23]} were fast and accurate landmark detectors, and provided robust capability of object alignment. For object tracking tasks, much more improvements have been made by correlation filters based methods. Minimum output sum of squared error (MOSSE) filter\cite{[2]} was the first correlation filter applied to tracking, which achieved good accuracy and real-time speed. Using multi-channel features or conducting feature fusion can improve the performance of the tracker\cite{re2_2,re2_1,re2_4,re2_3,re2_5}, in \cite{[kcf]}, kernelized correlation filter(KCF) used kernelized HOG feature to cope with texture variations, and utilized the properties of circulant matrices to obtain a fast solution in frequency domain. Some algorithms improved KCF by adding scale estimation\cite{DSST,Ding2017Real}, in \cite{DSST}, Danelljan et al. learned discriminative correlation filters based on a scale pyramid representation. In \cite{cf1}, Liu et al. introduced a structural correlation filter which takes part-based tracking into correlation filters to solve the tracking problem, and similar idea was also used in \cite{[54]}.
 
In general, both constrained and unconstrained correlation filtering algorithms work pretty well in ideal situations. However, in our complex industrial setting, the performance degrades dramatically when dealing with distorted data due to occlusion, noise and shifting. To balance efficiency with effectiveness of industrial applications, a reliable way of improving system performance is to add specific constraints to correlation filters. From this point of view, our work follows the constrained correlation filter strategy, where many similar methods have been proposed. In \cite{[boundary]}, a new correlation filter based on a limited boundary constraint could drastically reduce the number of examples affected by boundary effects in training process, which can significantly improve detection/tracking performance. The maximum margin correlation filters (MMCF) \cite{ccf1}, which constrain the output at the target location, demonstrating better robustness to outliers. The Distance Classifier Correlation Filter (DCCF) \cite{ccf2} incorporated the distance information into the filter calculation for multi-class tasks. In \cite{pcm}, an adaptive multi-class correlation filters (AMCF) method was developed based on an alternating direction method of multipliers (ADMM) framework by considering the multiple-class output information into the optimization objective.

Although all constrained correlation filter methods above can improve the performance to some extent with low-computational cost, they still suffer from less stable performance by various distorted patterns in industrial scenarios.
In correlation filtering algorithm, there is a basic fact that the solutions obtained by  different training samples are usually very different.  Training data containing various kinds of interference can produce robust correlation filters, contrarily, incomplete training data will make filters lose the robustness. We can explain this phenomenon as training data sampling(bagging) leading to solution sampling. Large amount of training data augumentation can obviously improve the robustness of the correlation filters, but will significantly increase computational cost and the risk of overfitting. We propose our Dijkstra-distance based correlation filters(DBCF) from another perspective, we think that the bagging results can be used to estimate the solution distribution, and this distribution, which is quantified by the shortest path among solutions, can be employed to improve the original solution. Actually, this is a compromise scheme, which can reduce the information loss caused by training data sampling, and avoid overfitting caused by large amount of data augumentation. In this paper, we aim to implement this idea in correlation filtering in order to enhance the robustness of the algorithm for dealing with various distorted patterns in complex industrial setting.

The overall framework of our proposed DBCF is shown in Fig. 1. Unlike an ad-hoc way that directly inputs all samples to train correlation filters, we train intermediate filters step by step based on iteratively selecting subsets. We build the reconstruction space by using the distribution information of these intermediate filters in metric space, and then project the solution into the reconstruction space to achieve optimization. Ultimately, we can find a better result from the subset that contains different kinds of solutions to our problem in various complex industrial settings.

\begin{figure*}
	\centering
	\includegraphics[width=0.98\linewidth]{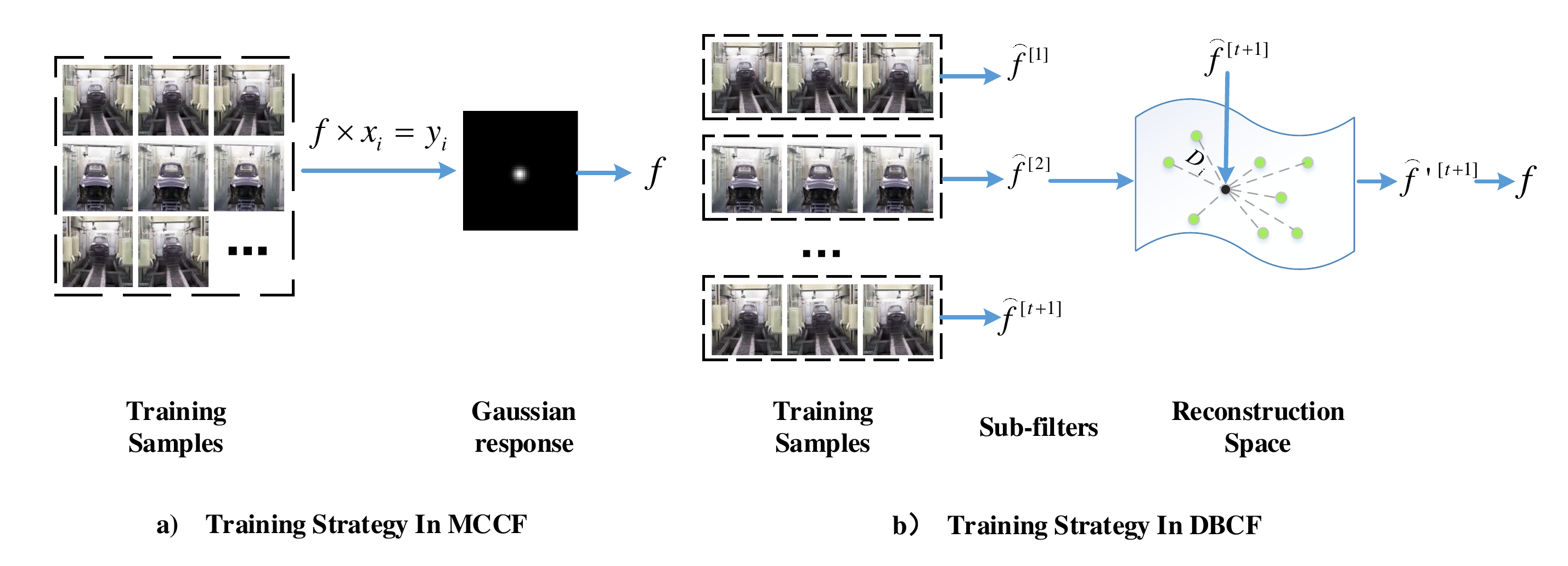}
	\caption{The framework of Dijkstra-distance based correlation filters. Different from MCCF that directly inputs all samples to train a filter $f$, we transform the original problem into solving a series of sub-filters($\hat{f}^{[0:t]}$), which are trained from different samples, then a reconstruction space is built. $\hat{f}^{[t+1]}$ and $\hat{f}'^{[t+1]} $ are both updated based on a iterative process, specifically, $\hat{f}^{[t+1]}$ is solved based on $\hat{f}'^{[t]}$, while $\hat{f}'^{[t+1]}$ is obtained based on a function $\Phi$ in the reconstruction space.}
	\label{fig:figframe}
\end{figure*}

Considering the time-critical aspect of industrial applications, another key issue of implementing such an idea is how to efficiently quantify distribution constraints. In this paper, we propose the Dijkstra-distance quantification. In summary, the main contributions of the proposed DBCF approach can be concluded in three aspects:

1) DBCF solves correlation filters in a reconstruction space based on the distribution-related constraints. During the training process, the new filters are corrected by the approximated distribution-related constrains step by step, where a stable correlation filter is eventually regularized.

2) To quantify the solution distribution, the shortest path among filters, named Dijkstra-distance, is used. The relationship between Euclidean distance and Dijkstra-space are also analyzed in our work.

3) A video dataset is built as a new benchmark for industrial 4.0 applications. These videos record the scene of automobile industry production line and object detection and tracking task are tested in this dataset.

\section{Dijkstra-distance Based Correlation Filter}
\label{sec:DBCf}
\subsection{Multi-channel correlation filters}

HOG and SIFT are descriptors which are widely used in object recognition. For MCCF, HOG descriptor is used to extract a multi-channel feature as input. Solving the problem of MCCF can be regarded as an optimization problem which is written as minimizing ${E}(f)$:

\begin{equation}
E({{f}}) = \frac{1}{2}\sum\limits_{i = 1}^N {||{{{y}}_i} - \sum\limits_{k = 1}^K {{{{f}}_k^{}}^T \otimes {{{x}}_{i,k}}} ||_2^2}  + \frac{\lambda}{2}\sum\limits_{k = 1}^K {||{{{f}}_k^{}}||_2^2},
\label{eq:eq1}
\end{equation}
where ${x}_i$ refers to a multiple channel input, ${y}_i$ is a single-channel Gaussian label whose peak is located at the center of target, $N$ represents the number of training samples, $\lambda$ is a regularization term, $K$ is the number of feature channels and $\otimes$ is cross correlation operator. Correlation filters are usually solved in the frequency domain because of the high efficiency, and therefore, we transform Eq.\ref{eq:eq1} into  frequency domain by the Fast Fourier Transform (FFT), which gives rise to:

\begin{equation}
E({{\hat f}}) = \frac{1}{2}\sum\limits_{i = 1}^N {||{{{{\hat y}}}_i} - {{{{\hat X}}}_i}} {{\hat f}}||_2^2 + \frac{\lambda }{2}||{{\hat f}}||_2^2,
\label{eq:eq2}
\end{equation}	
where $\hat{}$ refers to the Fourier form of the corresponding variable,  	${{\hat f}} = {[{{{{\hat f}}}_1^{T}},...,{{{{\hat f}}}_k^{T}}]^T}$,
${{\hat X}_i} = [{\rm{diag}}{({{\hat x}_{i,1}})^T},...,{\rm{diag}}{({{\hat x}_{i,k}})^T}]$.
A solution in the frequency domain is given by:
\begin{equation}
\hat{f} = {(\lambda {{I}} + \sum\limits_{i = 1}^N {{{{{\hat X}}}_i}^T{{{{\hat X}}}_i}})^{ - 1}}\sum\limits_{i = 1}^N {{{{{\hat X}}}_i}^T{{{{\hat y}}}_i}}.
\label{eq:eq3}
\end{equation}

\subsection{Dijkstra-distance}
\label{sec:Dijkstra}
In our DBCF, evaluating the distribution information of solutions (sub-filters) produced by different training samples is a crucial step for building reconstruction space, and using distance information is the most direct idea to evaluate this distribution.  Euclidean distance is the most commonly distance representation, whihc is widely used in distribution information evaluation\cite{Zhang2017Latent,[8]}. However, the solutions are always distributed in a high-dimensional space, the reconstruction space may not always be Euclidean, therefore, we need to extend  Euclidean distance to Dijkstra-distance. Dijkstra-distance is a classical single source shortest path algorithm, which uses a greedy strategy to calculate the shortest distance from source point to other points. Considering a point group with $N+1$ points, the pseudo code for solving the Dijkstra-distance $(D_{g})$ is shown as Algorithm.\ref{alg:dirjkstra}, which can further be summarized as Eq.\ref{eq:eq6}:
\begin{equation}
D_{g}(N_0,(N_1:N_n))=Dijkstra(N_0,(N_1:N_n),M)
\label{eq:eq6}
\end{equation}

In Eq.\ref{eq:eq6}, an important parameter is the number of neighborhood points ($M$). The value of $M$ determines whether two points in reconstruction space can be connected and which is the shortest path between them. Considering a ultimate situation, if $M$ is equal to the number of points in the reconstructed space, Dijkstra distance will be reduced to Euclidean distance $(D_{e})$ approximately, and the corresponding reconstructed space is also an approximate Euclidean space.

\begin{equation}
D_{e}(N_0,N_i)=||N_0-N_i||_{2}^{2}.
\label{eq:eq7}
\end{equation}

		\begin{algorithm}
			\caption{Dijkstra-distance Algorithm} 
			\label{alg:Dijkstra}
			\begin{algorithmic}[1]
				\State Set Source point:$N_0$, Other points: $N_1...N_n$
					\State Initialize $M$, $M$ refers to the number of neighbor points.
					\State Initialize the ergodic set: $A=\{N_0\}$ , point set: $B=\{N_1,N_2...N_n\}$
					\State Initialize Dijkstra-distance matrix $D_g=(d_{(0,1)},d_{(0,2)}....d_{(0,n)})$ (if $N_i$ is not the neighbour point of $N_0$, $d_{(0,i)}=+ \infty$)
					\State Initialize the current point $C=N_0$
					\Repeat	
					\State Find the nearest $N_i$ of $C$, set $V=N_i$
					\State Add $V$ to ergodic set: $A=A+V$
					\State Remove $V$ from point set: $B=B-V$
					\State Caluculate the distance between V and other points in $B$: $d_{(V,B_i)}$
					\State Update Dijkstra-distance matrix $D_g$:
					\If{$d_{(0,i)}>d_{(0,V)}+d_{(V,B_i)}$}
					\State $d_{(0,i)}=d_{(0,V)}+d_{(V,B_i)}$
					\EndIf
					\State Update $C=V$
					
					\Until{$B=\varnothing$}
					\State Output Dijkstra-distance matrix $D_g$
			\end{algorithmic}
			\label{alg:dirjkstra}
\end{algorithm}
\subsection{Dijkstra-distance Based Correlation Filter}
Traditional strategies can not provide satisfactory performance because the simple sampling(bagging) can not find the exact solutions from the whole various industrial data space. To obtain the exact solution,  large amount of data augmentation is needed to avoid data sampling, which will increase computational cost significantly. Based on MCCF, we explore a new strategy for learning a correlation filter. We derive a small number of training samples by adding noise or conducting affine transformation on original dataset step by step, and obtain several sub-filters (sampling solutions). These sub-filters will be used to form a reconstruction space latter. The purpose of doing so is to consider the distribution information of the sub-filters generated by different samples, further estimate a more accurate solution based on this reconstruction space. 
We assume that there is a mapping relationship between final solution and the reconstruction space, and set up a variable $f' $ as the projection of the final solution in this reconstruction space. Considering the constraints between $f$ and the reconstruction space, we rewrite the optimization problem described in Eq.\ref{eq:eq2} as:

\begin{equation}
\begin{aligned}
&E({{\hat f}}) = \frac{1}{2}\sum\limits_{i = 1}^N {||{{{{\hat y}}}_i} - {{{{\hat X}}}_i}} {{\hat f}}||_2^2 + \frac{\lambda }{2}||{{\hat f}}||_2^2,
\\&Subject\ to: \hat{f} \rightarrow \hat{f'}, \hat{f'}\in S
\end{aligned}
\label{eq:buchong1}
\end{equation}

As shown in Eq.\ref{eq:buchong1}, actually we add a constraint term to the original optimization problem. Specifically, we regard the filter $f$ as the variable to be sovled in the optimization problem, cloning a variable $f'$ as the map of $f$ in reconstruction subspace $S$, we rewrite Eq.\ref{eq:buchong1} in Lagrangian form as:

\begin{equation}
\begin{array}{l}
E(\hat{f}) = \frac{1}{2}\sum\limits_{i = 1}^{N} {||{{\hat{y}}_i} - {{\hat{X}}_i} \hat{f}} ||_2^2 + \frac{\lambda }{2}||\hat{f}||_2^2 +  \frac{\sigma}{2} ||\hat{f'} - \hat{f}|{|^2},  \\			
\end{array}
\label{eq:eq4}
\end{equation}
where $\lambda$ and $\sigma$ are regularization terms. The objective function is not easy to solve because of the additional constraint. As both $f$ and $f'$ are unknown variables, we can not obtain the close form solution of the optimization problem by the method of derivation. To this end, we introduce a method based on a iterative process which is similar to \cite{[8]}, the solution is given as follows:

\begin{equation}
\begin{aligned}
{{\hat{f}}^{[t + 1]}} &= {{{(\sum\limits_{i = 1}^N {({{\hat{X}}_i}^T{{\hat{X}}_i})}  + \lambda{I} + {\sigma^{[t]}}{I})^{-1}}}}\times(\sum\limits_{i = 1}^N {{{\hat{X}}_i}^T{{\hat{y}}_i}}  +  {\sigma^{[t]}}{\hat{f}'^{[t]}}),\\
{{\hat{f}}'^{[t + 1]}} &= \Phi ({{\hat{f}}^{[t+1]}}, {{\hat{f}}^{[0:t]}}) =
{{\sum\limits_{i = 1}^t {\omega_i}\hat{{f}}^{[i]}}}
=
\sum\limits_{i = 1}^t {Normalize}(\frac{1}{{D_i}})\hat{{f}}^{[i]}.
\end{aligned}
\label{eq:eq5}
\end{equation}
where $t$ is iteration number, $\Phi$ refers to a reconstruction function related to the distance of sub-filters, $D_i$ is the Dijkstra distance\cite{dijkstra1959note} between ${\hat{ f}}^{[t + 1]}$ and ${\hat{f}}^{[i]}$,
\begin{equation}
D_i=D_{g}(\hat{f}^{[t + 1]},\hat{f}^{[0:t]})=Dijkstra(\hat{f}^{[t + 1]},\hat{f}^{[0:t]},M)
\label{eq:buchong2}
\end{equation}
As described in Section\ref{sec:Dijkstra}, Euclidean distance is approximately regarded as a special case of Dijkstra distance. If we set $M=t+1$, $D_i$ will be calculated as Euclidean distance:
\begin{equation}
D_i=D_{e}(\hat{f}^{[t + 1]},\hat{f}^{[i]})=||\hat{f}^{[t + 1]}-\hat{f}^{[i]}||_{2}^{2}.
\label{eq:buchong3}
\end{equation}
 
As Eq.\ref{eq:eq5} implies, some sub-filters $(\hat{f}^{[0:t]})$ are produced during the iteration process, which will build so-called reconstruction space,  $\hat{f}'^{[t + 1]}$ is the projection of $\hat{f}^{[t + 1]}$ in this specific space, and is reconstructed from ${\hat{f}}^{[0:t]}$,
sub-filters with short distance are assigned with larger weights. After some iterations, $\hat{{f}}^{[t+1]}$ converges to a saddle point.

\subsection{Implementation}

As shown in Fig.\ref{fig:figframe}, compared with MCCF, we decompose the task of training a filter into training some sub-filters, establish the constraint by using distribution information among the filters, and obtain the final solution based on reconstruction space. The algorithm is solved through a iterative process, to ensure the convergence of algorithm, we set a threshold coefficient $\eta$ to control the penalty term $\sigma$. When the difference between $\hat{{f}}^{[t + 1]}$ and $\hat{{f}}^{[t]}$ dose not diminish, the value of $\sigma$ will be increased to accelerate the convergence process. The pseudocode of our proposed method is sumarized in  Algorithm.\ref{alg:dccf}.
\begin{algorithm}
	\caption{Dijkstra-distance Based Correlation Filters} \label{alg:dccf}
	\begin{algorithmic}[1]
		\State Set $t=0$, $\varepsilon_{best} = + \infty$, $\eta = 0.7$  (suggested in \cite{Zhang2017Latent})
		\State Initialize {$\lambda=10^{-4}$},$\sigma^{[0]} = 0.25$ (suggested in \cite{[kcf],[8]})
		\State Initialize $\hat{{f}}^{[0]}$ and $\hat{{f}}'^{[0]}$ based on MCCF
		\State Initialize $N=S_{0}$, $S_{t}$. $S_{0}$ denotes the size of half of training samples, $S_{t}$ represents the number of training samples during each iteration.
		\Repeat
		\State update ${\hat{f}^{[t + 1]}}$ using Eq.\ref{eq:eq5}
		\State $\varepsilon  = \|\hat{{f}}^{[t + 1]} - \hat{{f}}^{[t]} \|_2$
		\If{$\varepsilon  < \eta \times \varepsilon_{best}$}
		\State $\sigma^{[t+1]} = \sigma^{[t]}$
		\State $\varepsilon_{best} = \varepsilon$
		\Else
		\State $\sigma^{[t+1]} = 2 \sigma^{[t]}$
		\EndIf
		\State update ${\hat{f}'^{[t + 1]}}$ using Eq.\ref{eq:eq5} $D_{i}=D_e$ or $D_{i}=D_{g}$
		
		\State $t \leftarrow t+1$, $N \leftarrow N + S_{t}$
		\Until{some stopping criterion}
	\end{algorithmic}
\label{alg:alg1}
\end{algorithm}

\subsection{Computational and Memory Complexity}
Considering a D-dimensional vector features (i.e $\hat{X}_{i}$ in Eq.\ref{eq:eq2} and \ref{eq:eq4}), The major computational cost is from FFT operations, therefore, DBCF has a time cost of $\mathcal{O}(N DlogD)$, which is the same as MCCF. The memory storage is $\mathcal{O}(TD)$ for DBCF, which is not large because  $T$ (the number of sub-filters) is often set as a small integer.

\section{Experiments}

\label{sec:experiment}
In this section, to evaluate the performance of the proposed method, experiments are conducted for object detection and tracking in industrial setting. We collect twelve industrial videos and label them manually to build a new benchmark for quantitatively evaluating the detection and tracking results. These videos are filmed in automobile industry production line, whose resolution of images is 640*480. For industrial automation, two applications are considered: object detection and tracking. For all experiments, we choose the same parameter to generate the desired Gaussian output whose peak is located at the coordinate of target. All images are normalized before training and testing. The images are power normalized to have a zero-mean and a standard deviation of 1.0. 

\subsection{Detection}
\label{sec:detection}
In detection experiment, the proposed method is evaluated on an industrial detection benchmark we collected and compared with several state-of-the-art correlation filters in the literature including \cite{[2],[1],[3]}. Parameters of compared methods are optimized for best performances, which facilitates fair comparisons with ours.

\textbf{Dataset:} Different from natural scenes, industrial scene images usually contain complex lighting changes, noises and occlusions. We establish a database for industrial object detection, which contains 134 images. We double the database by flipping them and randomly select 200 images for training, coupled with the remainder 68 images for testing. 
All images are cropped to have a size from 150 $\times$ 200 to 200$\times$ 200 pixels.

\textbf{Parameter setting:} For ASEF\cite{[1]} and MOSSE\cite{[2]}, we extract intensity feature as they are single-channel correlation algorithm, and for MCCF and ours, we use HOG\cite{[4]}. The number of direction gradients is set to 5 and the sizes of block and cell are both [5,5] (as suggested in \cite{[3]}). For all of the algorithm, we define the required response(labels) as  a 2D Gaussian matrix with a variance of 2, and the peak is located at the center of the object to be detected. We use a $60 \times 60$ pixels filter for our object. Object detection is performed by correlating the filters over the testing images followed by selecting the peaks of the outputs as the predicted object locations.

\textbf{Training strategy:} For our DBCF, as described in Algorithm.\ref{alg:alg1}, we first select a subset which contains half of training samples $(S_0)$ to initialize $\hat{{f}}^{[0]}$ and $\hat{{f}}'^{[0]}$, and then we perform a process to generate other $T$ subsets by adding $S_{t}$ samples into initialized subset step by step. Then we can obtain $\hat{{f}}^{[t]}$ and $\hat{{f}}'^{[t]}$ according to the iterative process. The added $S_{t}$ samples can be original images or its variant version. If variants are applied, the reconstruction space will be more abundant and corresponding results will be more precise.
\begin{figure*}
	\centering
	\includegraphics[width=0.95\linewidth]{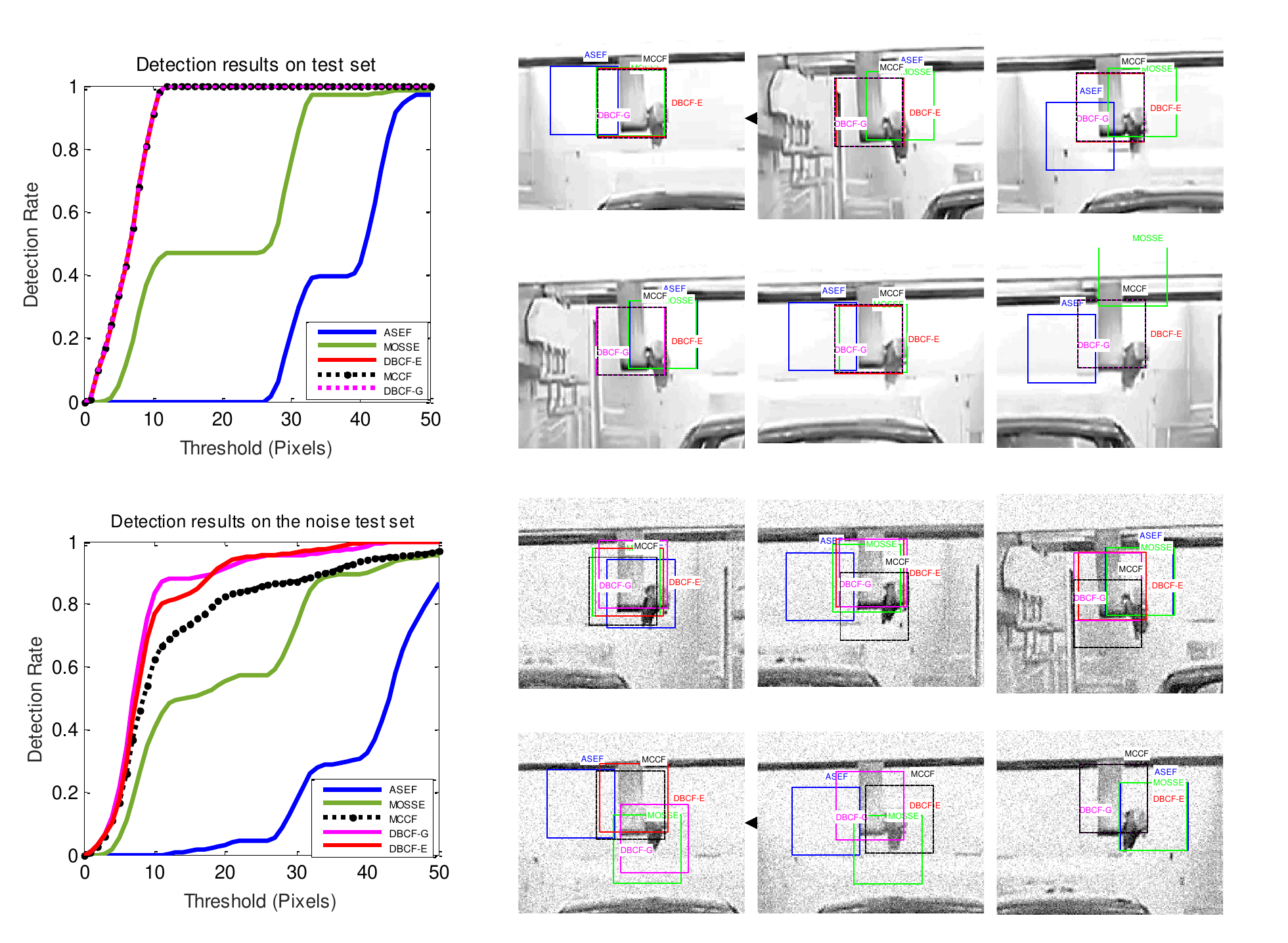}
	\caption{The results of DBCF compared to the state-of-the-art correlation filters on detection dataset. The upper picture group is illustrated for the original images, while the lower is illustrated for the  images suffering severe Gaussian noise(Best seen on screen.)}
	\label{fig:detection}
\end{figure*}

\textbf{Results:} For the detection experiment, we evaluate the performance of our algorithm by the distance of the predicted positions and groundtruth, which can be computed as:
\begin{equation}
d = {{||{p}_i} - {{m}_i}||_{2}^{2}}
\label{eq:eq8}
\end{equation}
where ${p}_i$ is the predicted location calculated by our method, and ${m}_i$ is the groundtruth of the target of interest. After calculating the distance $d$, the next step is to compare it with a threshold $\tau$. If $d < \tau$, the result will be considered as a correct one. We count the correct number under this threshold, and compute the ratio between this number and the number of test samples as the localization rate.

DBCF is compared with ASEF, MOSSE and MCCF in the robustness evaluation. As shown in Fig.~\ref{fig:detection}, on the original test set, ASEF and MOSSE are difficult to deal with complex variance in industrial environment, while MCCF and DBCF achieve a much higher performance due to the multi-channel feature. However, the accuracy of MCCF decreases with the increase of Gaussian noise obviously, while DBCF is less affected by noise than others. DBCF shows better robustness when the testing data suffers from heavy noise. This outstanding performance can be attributed to that we considering the distribution information of sampling solutions.  
\begin{figure}
	\includegraphics[width=0.85\textwidth]{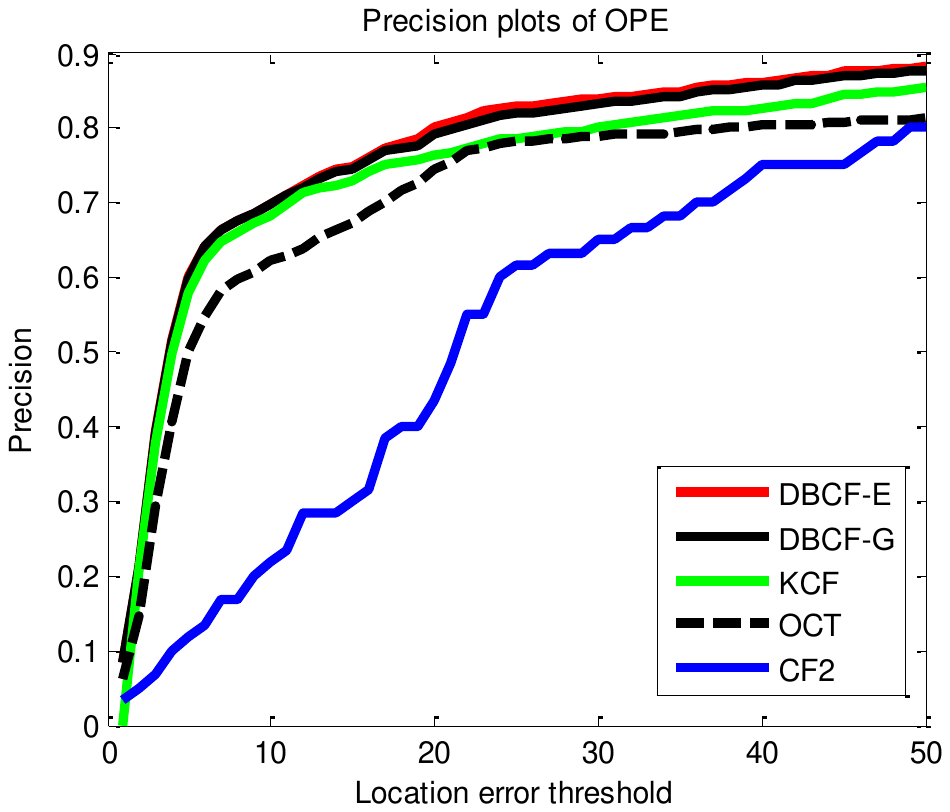}
	\caption{ Success and precision plots according to the online tracking benchmark.}
	\label{fig:precision}
\end{figure}
In these experiments, all methods are based on the same training and testing sets. A 10-fold cross validation procedure is employed to compute evaluation results.

\subsection{Tracking}
\label{sec:tracking}

The evaluation of object tracking with the proposed method is conducted on 12 sequences of the database we built. The Gaussian kernel function (standard variance = $0.5$) and most parameters used in DBCF are empirically chosen according to \cite{[kcf]}: $\lambda =10^{-4}$, $\rho=0.1$, and the searching size is set to 1.5.

For tracking experiment, we compare our method with several state-of-the-art algorithm\cite{[kcf]}, \cite{oct}, \cite{[cf2]}. Distance precision\cite{[24],[kcf],[cf2],[29],Zhang2017Adaptive,Zhang2016Bounding} is used as an evaluation criterion in this paper. Distance precision measure the ratio of successful tracking frames whose tracking deviation is within the given threshold. In Fig.\ref{fig:precision}, the overall success and precision plots of our DBCF and other contrast algorithms are reported. In terms of precision, DBCF and KCF respectively achieve 80.2\% and 76.4\% when the threshold is set to 20. Compared with OCT\cite{oct}, which is the one of latest variants of KCF, DBCF achieves a significant performance improvement in terms of precision (1\% improvement). We also compare with convolutional feature correlation filter(CF2)\cite{[cf2]}, a deep feature based method. Likewise, this deep feature does not show a strong discrimination in this industrial scene tracking benchmark and DBCF can achieve a better result with faster speed. Fig.4 shows some keys frames of the methods above during tracking process. These results confirm that the Dijkstra-distance constraint contributes to our tracker and enables it to perform better than the state-of-the-art trackers.
	
On an Intel I7 2.4 GZ (4 cores) CPU and 4G RAM, DBCF can run up to 129 FPS, while the KCF, OCT and CF2 achieve 210, 54, 1.1 FPS respectively. Without losing the real-time performance, the tracking performance is significantly improved by DBCF about 3.8\% on the precision.The training and testing sets and the source code is publicly available on github.

\begin{figure} [t!]
	\includegraphics[width=0.85\textwidth]{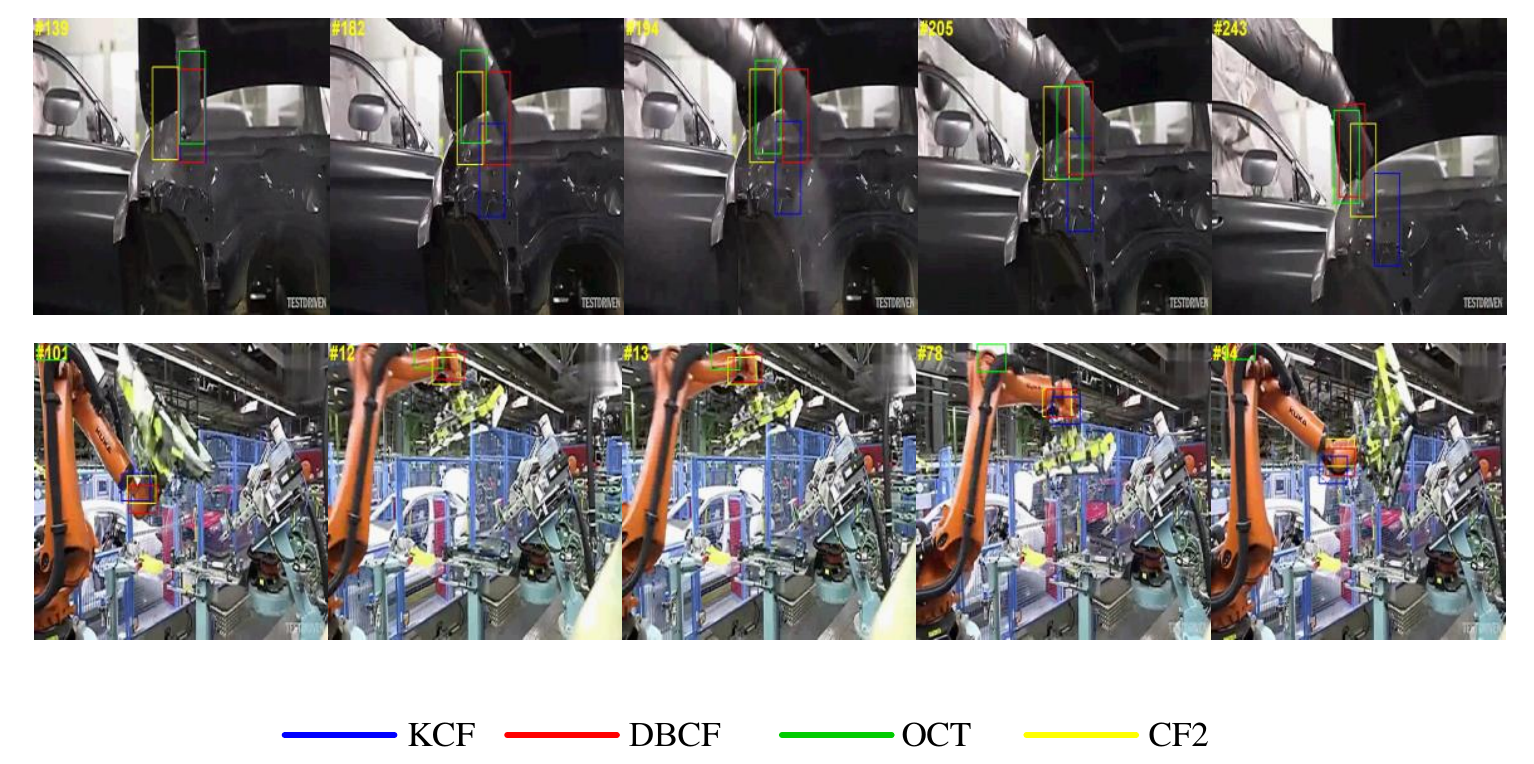}
	\caption{Illustration of some tracking key frames.}
\end{figure}

\section{Conclusions}
In this paper, we have proposed a Dijkstra-distance based correlation filters, named DBCF, in which the shortest path has been introduced to quantify the solution distribution. The theoretical analysis reveals that Dijksra-distance, as a generalization of Euclidean distance, is an effective way of distribution quantification. The experimental results have shown consistent advantages over the state-of-the-arts when applying DBCF to the industrial 4.0 applications of object detection and tracking. The future work will focus on action recognition\cite{Zhang2017Action} and image restoration\cite{onetwoone}.

\section{Acknowledgement}
The work was supported by the Natural Science Foundation of China under Contract 61672079 and 61473086, the Open Projects Program of National Laboratory of Pattern Recognition, and Shenzhen Peacock Plan KQTD20161-12515134654. Shangzhen Luan and Yan Li have the same contribution to this paper. We also want to thank Professor Zhu Lei of Hangzhou Dianzi University, who participated in writing and technical editing of the manuscript.


\begin{thebibliography}{10}

\bibitem{2008Real}
Selim Benhimane, Hesam Najafi, Matthias Grundmann, Ezio Malis, Yakup Genc, and
  Nassir Navab.
\newblock Real-time object detection and tracking for industrial applications.
\newblock In {\em Visapp 2008: Proceedings of the Third International
  Conference on Computer Vision Theory and Applications, Funchal, Madeira,
  Portugal, January}, pages 337--345, 2008.

\bibitem{[2]}
David~S Bolme, J~Ross Beveridge, Bruce~A Draper, and Yui~Man Lui.
\newblock Visual object tracking using adaptive correlation filters.
\newblock In {\em Computer Vision and Pattern Recognition (CVPR), 2010 IEEE
  Conference on}, pages 2544--2550. IEEE, 2010.

\bibitem{[1]}
David~S Bolme, Bruce~A Draper, and J~Ross Beveridge.
\newblock Average of synthetic exact filters.
\newblock In {\em Computer Vision and Pattern Recognition, 2009. CVPR 2009.
  IEEE Conference on}, pages 2105--2112. IEEE, 2009.

\bibitem{[4]}
Navneet Dalal and Bill Triggs.
\newblock Histograms of oriented gradients for human detection.
\newblock In {\em Computer Vision and Pattern Recognition, 2005. CVPR 2005.
  IEEE Computer Society Conference on}, volume~1, pages 886--893. IEEE, 2005.

\bibitem{DSST}
Martin Danelljan, Gustav H{\"a}ger, Fahad Khan, and Michael Felsberg.
\newblock Accurate scale estimation for robust visual tracking.
\newblock In {\em British Machine Vision Conference, Nottingham, September 1-5,
  2014}. BMVA Press, 2014.

\bibitem{dijkstra1959note}
Edsger~W Dijkstra.
\newblock A note on two problems in connexion with graphs.
\newblock {\em Numerische mathematik}, 1(1):269--271, 1959.

\bibitem{Ding2017Real}
Guiguang Ding, Wenshuo Chen, Sicheng Zhao, Jungong Han, and Qiaoyan Liu.
\newblock Real-time scalable visual tracking via quadrangle kernelized
  correlation filters.
\newblock {\em IEEE Transactions on Intelligent Transportation Systems},
  19(1):140--150, 2017.

\bibitem{Han2012Employing}
Jungong Han, Eric~J. Pauwels, Paul M.~De Zeeuw, and Peter H. N.~De With.
\newblock Employing a rgb-d sensor for real-time tracking of humans across
  multiple re-entries in a smart environment.
\newblock {\em IEEE Transactions on Consumer Electronics}, 58(2):255--263,
  2012.

\bibitem{[24]}
Joao Henriques, Rui Caseiro, Pedro Martins, and Jorge Batista.
\newblock Exploiting the circulant structure of tracking-by-detection with
  kernels.
\newblock {\em Computer Vision--ECCV 2012}, pages 702--715, 2012.

\bibitem{[kcf]}
Jo{\~a}o~F Henriques, Rui Caseiro, Pedro Martins, and Jorge Batista.
\newblock High-speed tracking with kernelized correlation filters.
\newblock {\em IEEE Transactions on Pattern Analysis and Machine Intelligence},
  37(3):583--596.

\bibitem{[9]}
Charles~F Hester and David Casasent.
\newblock Multivariant technique for multiclass pattern recognition.
\newblock {\em Applied Optics}, 19(11):1758--1761, 1980.

\bibitem{[15]}
Hamed Kiani, Terence Sim, and Simon Lucey.
\newblock Multi-channel correlation filters for human action recognition.
\newblock In {\em Image Processing (ICIP), 2014 IEEE International Conference
  on}, pages 1485--1489. IEEE, 2014.

\bibitem{[3]}
Hamed Kiani~Galoogahi, Terence Sim, and Simon Lucey.
\newblock Multi-channel correlation filters.
\newblock In {\em Proceedings of the IEEE International Conference on Computer
  Vision}, pages 3072--3079, 2013.

\bibitem{[boundary]}
Hamed Kiani~Galoogahi, Terence Sim, and Simon Lucey.
\newblock Correlation filters with limited boundaries.
\newblock In {\em Proceedings of the IEEE Conference on Computer Vision and
  Pattern Recognition}, pages 4630--4638, 2015.

\bibitem{re2_2}
X.~Lan, A.~J. Ma, P.~C. Yuen, and R~Chellappa.
\newblock Joint sparse representation and robust feature-level fusion for
  multi-cue visual tracking.
\newblock {\em IEEE Transactions on Image Processing A Publication of the IEEE
  Signal Processing Society}, 24(12):5826, 2015.

\bibitem{re2_1}
Xiangyuan Lan, A.~J Ma, and Pong~Chi Yuen.
\newblock Multi-cue visual tracking using robust feature-level fusion based on
  joint sparse representation.
\newblock In {\em Computer Vision and Pattern Recognition}, pages 1194--1201,
  2014.

\bibitem{re2_4}
Xiangyuan Lan, Pong~C Yuen, and Rama Chellappa.
\newblock Robust mil-based feature template learning for object tracking.
\newblock In {\em AAAI}, pages 4118--4125, 2017.

\bibitem{re2_3}
Xiangyuan Lan, Shengping Zhang, and Pong~C. Yuen.
\newblock Robust joint discriminative feature learning for visual tracking.
\newblock In {\em International Joint Conference on Artificial Intelligence},
  pages 3403--3410, 2016.

\bibitem{re2_5}
Xiangyuan Lan, Shengping Zhang, Pong~C. Yuen, and Rama Chellappa.
\newblock Learning common and feature-specific patterns: A novel
  multiple-sparse-representation-based tracker.
\newblock {\em IEEE Transactions on Image Processing}, PP(99):1--1, 2017.

\bibitem{cf1}
Si~Liu, Tianzhu Zhang, Xiaochun Cao, and Changsheng Xu.
\newblock Structural correlation filter for robust visual tracking.
\newblock In {\em Proceedings of the IEEE Conference on Computer Vision and
  Pattern Recognition}, pages 4312--4320, 2016.

\bibitem{[54]}
Ting Liu, Gang Wang, and Qingxiong Yang.
\newblock Real-time part-based visual tracking via adaptive correlation
  filters.
\newblock In {\em Proceedings of the IEEE Conference on Computer Vision and
  Pattern Recognition}, pages 4902--4912, 2015.

\bibitem{[cf2]}
Chao Ma, Jia-Bin Huang, Xiaokang Yang, and Ming-Hsuan Yang.
\newblock Hierarchical convolutional features for visual tracking.
\newblock In {\em Proceedings of the IEEE International Conference on Computer
  Vision}, 2015.

\bibitem{[29]}
Chao Ma, Xiaokang Yang, Chongyang Zhang, and Ming-Hsuan Yang.
\newblock Long-term correlation tracking.
\newblock In {\em Proceedings of the IEEE Conference on Computer Vision and
  Pattern Recognition}, pages 5388--5396, 2015.

\bibitem{[10]}
Abhijit Mahalanobis, BVK~Vijaya Kumar, and David Casasent.
\newblock Minimum average correlation energy filters.
\newblock {\em Applied Optics}, 26(17):3633--3640, 1987.

\bibitem{ccf2}
Abhijit Mahalanobis, BVK~Vijaya Kumar, and SRF Sims.
\newblock Distance-classifier correlation filters for multiclass target
  recognition.
\newblock {\em Applied Optics}, 35(17):3127--3133, 1996.

\bibitem{[13]}
Abhijit Mahalanobis, BVK~Vijaya Kumar, Sewoong Song, SRF Sims, and JF~Epperson.
\newblock Unconstrained correlation filters.
\newblock {\em Applied Optics}, 33(17):3751--3759, 1994.

\bibitem{[23]}
Vishnu Naresh~Boddeti, Takeo Kanade, and BVK Vijaya~Kumar.
\newblock Correlation filters for object alignment.
\newblock In {\em Proceedings of the IEEE Conference on Computer Vision and
  Pattern Recognition}, pages 2291--2298, 2013.

\bibitem{[11]}
Ph~Refregier.
\newblock Optimal trade-off filters for noise robustness, sharpness of the
  correlation peak, and horner efficiency.
\newblock {\em Optics Letters}, 16(11):829--831, 1991.

\bibitem{ccf1}
Andres Rodriguez, Vishnu~Naresh Boddeti, BVK~Vijaya Kumar, and Abhijit
  Mahalanobis.
\newblock Maximum margin correlation filter: A new approach for localization
  and classification.
\newblock {\em IEEE Transactions on Image Processing}, 22(2):631--643, 2013.

\bibitem{[12]}
Marios Savvides and BVK~Vijaya Kumar.
\newblock Efficient design of advanced correlation filters for robust
  distortion-tolerant face recognition.
\newblock In {\em Advanced Video and Signal Based Surveillance, 2003.
  Proceedings. IEEE Conference on}, pages 45--52. IEEE, 2003.

\bibitem{pcm}
Linlin Yang, Chen Chen, Hainan Wang, Baochang Zhang, and Jungong Han.
\newblock Adaptive multi-class correlation filters.
\newblock In {\em Pacific Rim Conference on Multimedia}, pages 680--688.
  Springer, 2016.

\bibitem{onetwoone}
Baochang Zhang, Jiaxin Gu, Chen Chen, Jungong Han, Xiangbo Su, Xianbin Cao, and
  Jianzhuang Liu.
\newblock One-two-one network for compression artifacts reduction in remote
  sensing.
\newblock {\em ISPRS Journal of Photogrammetry and Remote Sensing}, 2018.

\bibitem{oct}
Baochang Zhang, Zhigang Li, Xianbin Cao, Qixiang Ye, Chen Chen, Linlin Shen,
  Alessandro Perina, and Rongrong Jill.
\newblock Output constraint transfer for kernelized correlation filter in
  tracking.
\newblock {\em IEEE Transactions on Systems Man and Cybernetics Systems},
  47(4):693--703, 2016.

\bibitem{Zhang2017Adaptive}
Baochang Zhang, Zhigang Li, Alessandro Perina, Alessio~Del Bue, Vittorio
  Murino, and Jianzhuang Liu.
\newblock Adaptive local movement modeling for robust object tracking.
\newblock {\em IEEE Transactions on Circuits and Systems for Video Technology},
  27(7):1515--1526, 2017.

\bibitem{Zhang2017Latent}
Baochang Zhang, Shangzhen Luan, Chen Chen, Jungong Han, Wei Wang, Alessandro
  Perina, and Ling Shao.
\newblock Latent constrained correlation filter.
\newblock {\em IEEE Transactions on Image Processing}, PP(99):1--1, 2017.

\bibitem{Zhang2016Bounding}
Baochang Zhang, Alessandro Perina, Zhigang Li, Vittorio Murino, Jianzhuang Liu,
  and Rongrong Ji.
\newblock Bounding multiple gaussians uncertainty with application to object
  tracking.
\newblock {\em International Journal of Computer Vision}, 118(3):364--379,
  2016.

\bibitem{[8]}
Baochang Zhang, Alessandro Perina, Vittorio Murino, and Alessio Del~Bue.
\newblock Sparse representation classification with manifold constraints
  transfer.
\newblock In {\em Proceedings of the IEEE conference on computer vision and
  pattern recognition}, pages 4557--4565, 2015.

\bibitem{Zhang2017Action}
Baochang Zhang, Yun Yang, Chen Chen, Linlin Yang, Jungong Han, and Ling Shao.
\newblock Action recognition using 3d histograms of texture and a multi-class
  boosting classifier.
\newblock {\em IEEE Transactions on Image Processing}, 26(10):4648--4660, 2017.

\end{thebibliography}

\end{document}